\documentclass[11pt,a4paper]{article}
\usepackage[hyperref]{naaclhlt2018}
\usepackage{times}
\usepackage{latexsym}

\usepackage{url}

\usepackage{graphicx}  
\usepackage{amsmath}
\usepackage{amssymb}
\usepackage{xspace}
\usepackage{multirow}
\DeclareMathOperator*{\argmax}{arg\!max}

\usepackage{algorithm, algorithmicx, algpseudocode}
\usepackage{makecell}
\usepackage{array}
\usepackage{multirow}

\aclfinalcopy 

\newcommand{\tabincell}[2]{\begin{tabular}{@{}#1@{}}#2\end{tabular}}

\title{Generative Bridging Network for Neural Sequence Prediction}

\author{Wenhu Chen,$^1$ Guanlin Li,$^2$ Shuo Ren,$^5$ Shujie Liu,$^3$ Zhirui Zhang,$^4$ Mu Li,$^3$ Ming Zhou$^3$\\ 
University of California, Santa Barbara$^1$\\
Harbin Institute of Technology$^2$\\
Microsoft Research Asia$^3$\\
University of Science and Technology of China$^4$\\
Beijing University of Aeronautics and Astronautics$^5$\\
\small
wenhuchen@cs.ucsb.edu epsilonlee.green@gmail.com \{v-shure, shujliu, v-zhirzh, muli, mingzhou\}@microsoft.com \\
}

\date{}

\begin{document}

\newcommand{\learner}{\ensuremath{p_{\theta}(Y|X)}\xspace}
\newcommand{\learnert}{\ensuremath{p_{\theta}(y_t|y_{1:t-1},X)}\xspace}
\newcommand{\bridge}{\ensuremath{p_{\eta}(Y|Y^*)}\xspace}
\newcommand{\bridget}{\ensuremath{p_{\eta}(y_t|y_{1:t-1},Y^*)}\xspace}
\newcommand{\payoff}{\ensuremath{q(Y|Y^*)}\xspace}
\newcommand{\lossb}{\ensuremath{L_B(\eta)}\xspace}
\newcommand{\lossg}{\ensuremath{L_G(\theta)}\xspace}
\newcommand{\reward}{\ensuremath{R(Y|Y^*)}\xspace}
\newcommand{\rewardt}{\ensuremath{s(y_t|y_{1:t-1},Y^*)}\xspace}
\newcommand{\expect}[1]{\ensuremath{\underset{#1}{\mathbb{E}}}\xspace}
\newcommand{\union}[1]{\ensuremath{\underset{#1}{\cup}}\xspace}
\newcommand{\summation}[1]{\sum_{#1}\xspace}
\newcommand{\algorithmautorefname}{Algorithm}
\newcommand\StateX{\Statex\hspace{\algorithmicindent}}
\newcommand{\similaritytau}{\frac{S(Y, Y^*)}{\tau}}

\maketitle
\begin{abstract}
In order to alleviate data sparsity and overfitting problems in maximum likelihood estimation (MLE) for sequence prediction tasks, we propose the Generative Bridging Network (GBN), in which a novel bridge module is introduced to assist the training of the sequence prediction model (the generator network). Unlike MLE directly maximizing the conditional likelihood, the bridge extends the point-wise ground truth to a bridge distribution conditioned on it, and the generator is optimized to minimize their KL-divergence. 
Three different GBNs, namely uniform GBN, language-model GBN and coaching GBN, are proposed to penalize confidence, enhance language smoothness and relieve learning burden. Experiments conducted on two recognized sequence prediction tasks (machine translation and abstractive text summarization) show that our proposed GBNs can yield significant improvements over strong baselines. Furthermore, by analyzing samples drawn from different bridges, expected influences on the generator are verified.
\end{abstract}

\section{Introduction}
Sequence prediction has been widely used in tasks where the outputs are sequentially structured and mutually dependent. Recently, massive explorations in this area have been made to solve practical problems, such as machine translation ~\cite{bahdanau2014neural,ma2017softmax,norouzi2016reward}, syntactic parsing~\cite{vinyals2015grammar}, spelling correction~\cite{bahdanau2014neural}, image captioning~\cite{xu2015show} and speech recognition~\cite{chorowski2015attention}. Armed with modern computation power, deep LSTM~\cite{hochreiter1997long} or GRU~\cite{chung2014empirical} based
neural sequence prediction models have achieved the state-of-the-art performance. 

\begin{figure}[!t]
\centering
\includegraphics[width=1.0\linewidth]{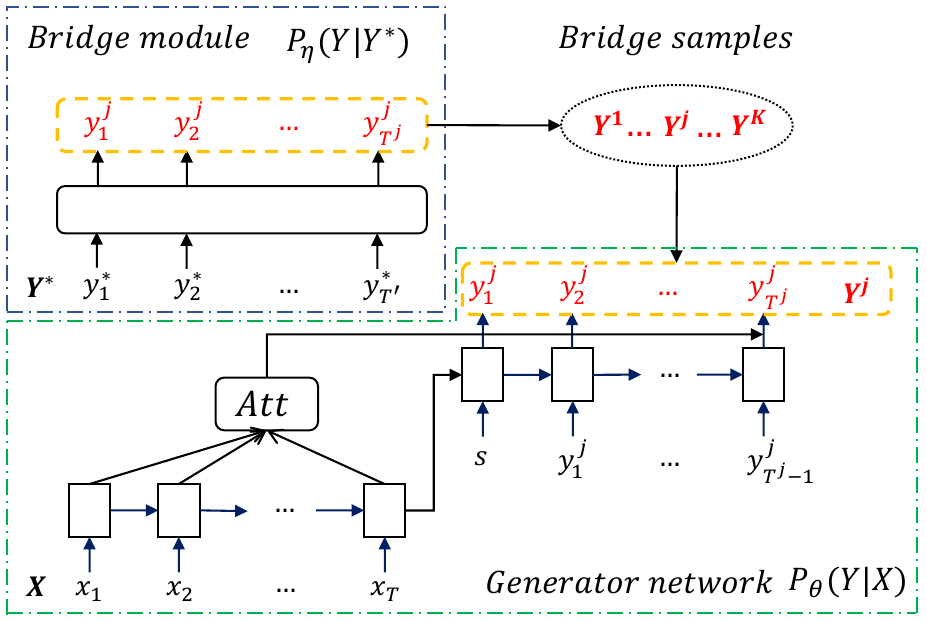}
\caption{The overall architecture of our novel Generative Bridging Network (GBN). Two main components, namely the generator network and the bridge module, are connected through samples ($Y^1 \dots Y^K$ in red) from the bridge module during training time. (We sometimes call them generator and bridge in brief respectively in the following discussion.) The generator is implemented through an attentive encoder-decoder, where in the figure \textit{Att} represents the attention module.}
\label{fig:GBN_arch}
\end{figure}

The typical training algorithm for sequence prediction is Maximum Likelihood Estimation (MLE), which  maximizes the likelihood of the target sequences conditioned on the source ones:
\begin{align}
\small
\begin{split}
\theta^* = \argmax_\theta \expect{(X,Y^*) \sim D} \log{p_{\theta}(Y^*|X)}
\end{split}
\end{align}
Despite the popularity of MLE or teacher forcing~\cite{doya1992bifurcations} in neural sequence prediction tasks, two general issues are always haunting: 1). data sparsity and 2). tendency for overfitting, with which can both harm model generalization. 

To combat data sparsity, different strategies have been proposed. Most of them try to take advantage of monolingual data~\cite{sennrich2015improving, zhang2016exploiting, cheng2016semi}.
Others try to modify the ground truth target based on derived rules to get more similar examples for training~\cite{norouzi2016reward, ma2017softmax}.
To alleviate overfitting, regularization techniques, such as confidence penalization~\cite{pereyra2017regularizing} and posterior regularization~\cite{zhang2017prior},  are proposed recently. 
As shown in Figure \ref{fig:GBN_arch}, we propose a novel learning architecture, titled Generative Bridging Network (GBN), to combine both of the benefits from synthetic data and regularization. Within the architecture, the bridge module (bridge) first transforms the point-wise ground truth into a bridge distribution, which can be viewed as a target proposer from whom more target examples are drawn to train the generator. 
By introducing different constraints, the bridge can be set or trained to possess specific property, with which the drawn samples can augment target-side data (alleviate data sparsity) while regularizing the training (avoid overfitting) of the generator network (generator).

In this paper, we introduce three different constraints to build three bridge modules. Together with the generator network, three GBN systems are constructed: 1). a uniform GBN, instantiating the constraint as a uniform distribution to penalize confidence; 2). a language-model GBN, instantiating the constraint as a pre-trained neural language model to increase language smoothness; 3). a coaching GBN, instantiating the constraint as the generator's output distribution to seek a close-to-generator distribution, which enables the bridge to draw easy-to-learn samples for the generator to learn. 
Without any constraint, our GBN degrades to MLE. The uniform GBN is proved to minimize KL-divergence with a so-called payoff distribution as in reward augmented maximum likelihood or RAML~\cite{norouzi2016reward}. 

Experiments are conducted on two sequence prediction tasks, namely machine translation and abstractive text summarization. On both of them, our proposed GBNs can significantly improve task performance, compared with strong baselines. Among them, the coaching GBN achieves the best. Samples from these three different bridges are demonstrated to confirm the expected impacts they have on the training of the generator. 
In summary, our contributions are:
\begin{itemize}
\item A novel GBN architecture is proposed for sequence prediction to alleviate the data sparsity and overfitting problems, where the bridge module and the generator network are integrated and jointly trained. 
\item Different constraints are introduced to build GBN variants: uniform GBN, language-model GBN and coaching GBN. Our GBN architecture is proved to be a generalized form of both MLE and RAML.
\item  All proposed GBN variants outperform the MLE baselines on machine translation and abstractive text summarization. Similar relative improvements are achieved compared to recent state-of-the-art methods in the translation task. We also demonstrate the advantage of our GBNs qualitatively by comparing ground truth and samples from bridges.
\end{itemize}

\begin{figure}[htb]
\centering
\includegraphics[width=1.0\linewidth]{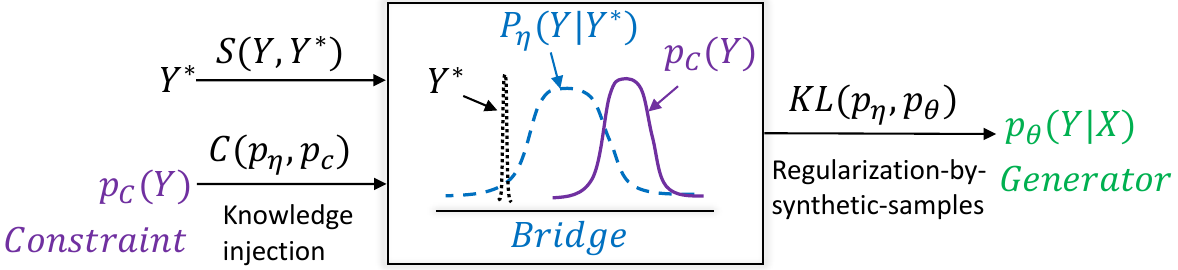}
\caption{Conceptual interpretation of our Generative Bridging Network (GBN). See detailed discussion in the beginning of Sec. \ref{sec.gbn}.}
\label{conceptual_interpret}
\end{figure}

\section{Generative Bridging Network}
\label{sec.gbn}
In this section, we first give a conceptual interpretation of our novel learning architecture which is sketched in Figure \ref{conceptual_interpret}. Since data augmentation and regularization are two golden solutions for tackling data sparsity and overfitting issues. We are willing to design an architecture which can integrate both of their benefits. The basic idea is to use a so-called bridge which transforms $Y^*$ to an easy-to-sample distribution, and then use this distribution (samples) to train and meanwhile regularize the sequence prediction model (the generator). 

The bridge is viewed as a conditional distribution\footnote{$\eta$ should be treated as an index of the bridge distribution, so it is not necessarily the parameters to be learned.} $\bridge$ to get more target $Y$s given $Y^*$ so as to construct more training pairs $(X, Y)$. In the meantime, we could inject (empirical) prior knowledge into the bridge through its optimization objective which is inspired by the design of the payoff distribution in RAML. We formulate the optimization objective with two parts in Equation (\ref{bridge_obj_raw}): a) an expected similarity score computed through a similarity score function $S(\cdot, Y^*)$ interpolated with b) a knowledge injection constraint\footnote{Note that, in our paper, we specify $\mathcal{C}$ to be KL-divergence between the bridge distribution $p_\eta$ and certain constraint distribution $p_c$, however, we believe mathematical form of $\mathcal{C}$ is not restricted, which could motivate further development.} $\mathcal{C}(\bridge, p_c(Y))$ where $\alpha$ controls the strength of the regularization, formally, we write the objective function $\lossb$ as follows: 
\begin{align}
\small
\begin{split}
\lossb = \expect{Y \sim p_{\eta}} [-S(Y,Y^*)] + \alpha \mathcal{C}(\bridge, p_c(Y))
\end{split}
\label{bridge_obj_raw}
\end{align}
Minimizing it empowers the bridge distribution not only to concentrate its mass around the ground truth $Y^*$ but also to adopt certain hope property from $p_c(Y)$.
With the constructed bridge distribution, we optimize the generator network $ P_\theta(Y \vert X) $ to match its output distribution towards the bridge distribution by minimizing their KL-divergence:
\begin{equation}
\small
\lossg = KL(\bridge \vert \vert \learner)
\label{generator_obj}
\end{equation}
In practice, the KL-divergence is approximated through sampling process detailed in Sec. \ref{sec.training}. As a matter of fact, the bridge is the crux of the integration: it synthesizes new targets to alleviate data sparsity and then uses the synthetic data as regularization to overcome overfitting. Thus a regularization-by-synthetic-example approach, which is very similar to the prior-incorporation-by-virtual-example method~\cite{niyogi1998incorporating}.

\subsection{Generator Network}
Our generator network is parameterized with the commonly used encoder-decoder architecture~\cite{bahdanau2014neural,cho2014learning}. The encoder is used to encode the input sequence $X$ to a sequence of hidden states, based on which an attention mechanism is leveraged to compute context vectors at the decoding stage. The context vector together with previous decoder's hidden state and previously predicted label are used, at each time step, to compute the next hidden state and predict an output label.

As claimed in Equation (\ref{generator_obj}), the generator network is not trained to maximize the likelihood of the ground truth but tries best to match the bridge distribution, which is a delegate of the ground truth. We use gradient descent to optimize the KL-divergence with respect to the generator:
\begin{align}
\small
\begin{split}
\nabla{\lossg} = \expect{Y \sim \bridge} \nabla\log{\learner}
\end{split}
\label{eq:gen_update}
\end{align}
The optimization process can be viewed as the generator maximizing the likelihood of samples drawn from the bridge. This may alleviate data sparsity and overfitting by posing more unseen scenarios to the generator and may help the generator generalize better in test time.

\subsection{Bridge Module\footnote{Although we name it bridge module, we explicitly learn it with the generator when a closed-form static solution exists in terms of Equation (\ref{bridge_obj}). Otherwise, we will adopt an encoder-decoder to construct a dynamic bridge network.}}

Our bridge module is designed to transform a single target example $Y^*$ to a bridge distribution \bridge. We design its optimization target in Equation (\ref{bridge_obj_raw}) to consist of two terms, namely, a concentration requirement and a constraint. The constraint is instantiated as KL-divergence between the bridge and a contraint distribution $p_c(Y)$. We transform Equation (\ref{bridge_obj_raw}) as follows, which is convenient for mathematical manipulation later:
\begin{align}
\small
\begin{split}
&\lossb = \expect{Y \sim p_{\eta}} [- \frac{S(Y,Y^*)}{\tau}] + KL(\bridge \vert \vert p_c(Y))
\end{split}
\label{bridge_obj}
\end{align}
$S(Y,Y^*)$ is a predefined score function which measures similarity between $Y$ and $Y^*$ and peaks when $Y = Y^*$, while $p_c(Y)$ reshapes the bridge distribution. More specifically, the first term ensures that the bridge should concentrate around the ground truth $Y^*$, and the second introduces willing property which can help regularize the generator. The hyperparameter $\tau$ can be interpreted as a temperature which scales the score function. In the following bridge specifications, the score function $S(Y, Y^*)$ is instantiated according to Sec. \ref{sec.smf}. 

\paragraph{1. Delta Bridge}
The delta bridge can be seen as the simplest case where $\alpha = 0$ or no constraint is imposed. The bridge seeks to minimize $ \expect{Y \sim p_\eta(Y|Y^*)} [-\frac{S(Y,Y^*)}{\tau}]$. The optimal solution is when the bridge only samples $Y^*$, thus the Dirac delta distribution is described as follows:
\begin{align}
\small
\bridge = \delta_{Y^*}(Y)
\end{align}
This exactly corresponds to MLE, where only examples in the dataset are used to train the generator. We regard this case as our baseline.

\paragraph{2. Uniform Bridge}
The uniform bridge adopts a uniform distribution $U(Y)$ as constraint. This bridge motivates to include noise into target example, which is similar to label smoothing~\cite{szegedy2016rethinking}. The loss function can be written as:
\begin{align}
\small
\begin{split}
&\lossb = \expect{Y \sim p_{\eta}} [-\frac{S(Y,Y^*)}{\tau}] + KL(\bridge \vert \vert U(Y))
\end{split}
\label{uniform_bridge_obj}
\end{align}
We can re-write it as follows by adding a constant to not change the optimization result:
\begin{align}
\small
\begin{split}
\lossb + C = KL(\bridge || {\frac{\exp \frac{S(Y,Y^*)}
{\tau}}{Z}})
\end{split}
\label{uniform_bridge_obj_mod}
\end{align}
This bridge is static for having a closed-form solution:
\begin{align}
\small
p_\eta(Y|Y^*) = \frac{\exp{\frac{S(Y,Y^*)}{\tau}}}{Z}
\end{align}
where $Z$ is the partition function. Note that our uniform bridge corresponds to the payoff distribution described in RAML~\cite{norouzi2016reward}.

\paragraph{3. Language-model (LM) Bridge}
The LM bridge utilizes a pretrained neural language model $p_{LM}(Y)$ as constraint, which motivates to propose target examples conforming to language fluency.
\begin{align}
\small
\begin{split}
&\lossb = \expect{Y \sim p_{\eta}} [-\frac{S(Y,Y^*)}{\tau}] + KL(\bridge||p_{LM})
\end{split}
\label{lm_bridge_obj}
\end{align}
Similar to the uniform bridge case, we can re-write the loss function to a KL-divergence:
\begin{align}
\small
\begin{split}
\lossb &= KL( \bridge || \hat{p}(Y)) + C \\
\hat{p}(Y) &= \frac{p_{LM}(Y) \cdot \exp \frac{S(Y,Y^*)}{\tau}}{Z}
\end{split}
\label{lm_bridge_obj_mod}
\end{align}
Thus, the LM bridge is also static and can be seen as an extension of the uniform bridge, where the exponentiated similarity score is re-weighted by a pretrained LM score, and renormalized:
\begin{align}
\small
\begin{split}
p(Y|Y^*) = \frac{p_{LM}(Y)\exp{\frac{S(Y,Y^*)}{\tau}}}{Z}
\end{split}
\end{align}
where $Z$ is the partition function. The above equation looks just like the payoff distribution, whereas an additional factor is considered. 

\paragraph{4. Coaching Bridge}
The coaching bridge utilizes the generator's output distribution as constraint, which motivates to generate training samples which are easy to be understood by the generator, so as to relieve its learning burden. The coaching bridge follows the same spirit as the coach proposed in Imitation-via-Coaching~\cite{he2012imitation}, which, in reinforcement learning vocabulary, advocates to guide the policy (generator) with easy-to-learn action trajectories and let it gradually approach the oracle when the optimal action is hard to achieve.
\begin{align}
\small
 \begin{split}
 &\lossb = \expect{Y \sim p_{\eta}} [-\frac{S(Y,Y^*)}{\tau}]  + KL(\learner||\bridge) 
\end{split}
\label{coaching_bridge_obj}
\end{align}
Since the KL constraint is a moving target when the generator is updated, the coaching bridge should not remain static. Therefore, we perform iterative optimization to train the bridge and the generator jointly.
Formally, the derivatives for the coaching bridge are written as follows:
\begin{align}
\small
\begin{split}
\nabla \lossb = &\expect{Y \sim p_{\eta}}  [-\frac{S(Y,Y^*)}{\tau}\nabla \log{\bridge}]  + \\
&\expect{Y \sim p_{\theta}} [\nabla \log{\bridge}]
\end{split}
\label{bridge_update}
\end{align}
The first term corresponds to the policy gradient algorithm described in REINFORCE~\cite{williams1992simple}, where the coefficient $-S(Y, Y^*)/\tau$ corresponds to reward function. Due to the mutual dependence between bridge module and generator network, we design an iterative training strategy, i.e. the two networks take turns to update their own parameters treating the other as fixed.

\subsection{Training}
\label{sec.training}

\begin{figure}[!t]
\centering
\includegraphics[width=1.0\linewidth]{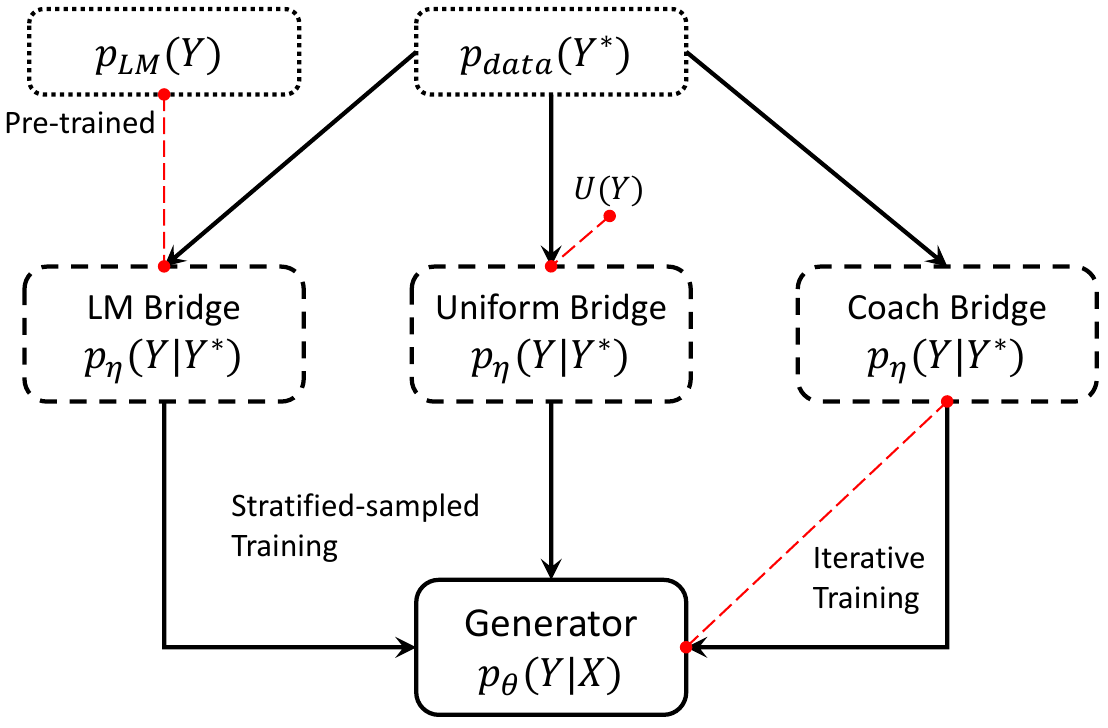}
\caption{The training processes of the three different variants of our GBN architecture (Sec.  \ref{sec.training}).}
\label{fig:Model-Architecture}
\end{figure}

The training of the above three variants is illustrated in Figure~\ref{fig:Model-Architecture}. Since the proposed bridges can be divided into static ones, which only require pre-training, and dynamic ones, which require continual training with the generator, we describe their training process in details respectively. 

\begin{figure}[tbh]
\centering
\includegraphics[width=1.0\linewidth]{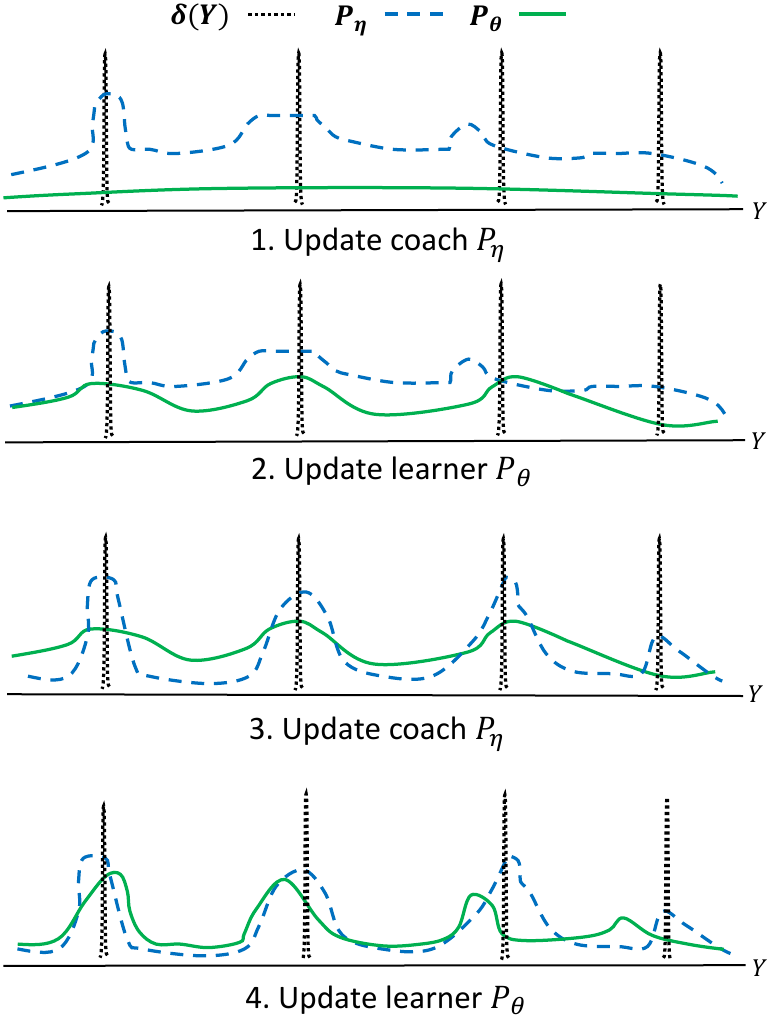}
\caption{Four iterative updates of the coaching bridge and the generator. In an early stage, the pre-trained generator $ P_\theta $ may not put mass on some ground truth target points within the output space, shown by $ \delta(Y) $. The coaching bridge is first updated with Equation (\ref{bridge_update}) to locate in between the Dirac delta distribution and the generator's output distribution. Then, by sampling from the coaching bridge for approximating Equation (\ref{eq:gen_update}), target samples which demonstrate easy-to-learn sequence segments facilitate the generator to be optimized to achieve closeness with the coaching bridge. Then this process repeats until the generator converges.}
\label{fig:Op-Diagram}
\end{figure}

\subsubsection{Stratified-Sampled Training}
Since closed-formed optimal distributions can be found for uniform/LM GBNs, we only need to draw samples from the static bridge distributions to train our sequence generator. Unfortunately, due to the intractability of these bridge distributions, direct sampling is infeasible. Therefore, we follow~\citet{norouzi2016reward, ma2017softmax} and adopt stratified sampling to approximate the direct sampling process. Given a sentence $Y^*$, we first sample an edit distance $m$, and then randomly select $m$ positions to replace the original tokens. The difference between the uniform and the LM bridge lies in that the uniform bridge replaces labels by drawing substitutions from a uniform distribution, while LM bridge takes the history as condition and draws substitutions from its step-wise distribution.

\subsubsection{Iterative Training}
Since the KL-constraint is a moving target for the coaching bridge, an iterative training strategy is designed to alternately update both the generator and the bridge (Algorithm~\ref{alg:coaching}). We first pre-train both the generator and the bridge and then start to alternately update their parameters. Figure~\ref{fig:Op-Diagram} intuitively demonstrates the intertwined optimization effects over the coaching bridge and the generator. We hypothesize that iterative training with easy-to-learn guidance could benefit gradient update, thus result in better local minimum. 

\begin{algorithm}[!t]
\caption{Training Coaching GBN}
\label{alg:coaching}
\begin{algorithmic}[0]
\Procedure{Pre-training}{}
\State Initialize \learner and \bridge with random weights $\theta$ and $\eta$ 
\State Pre-train \learner to predict $Y^*$ given X
\State Use pre-trained \learner to generate $\hat{Y}$ given X
\State Pre-train \bridge to predict $\hat{Y}$ given $Y^*$
\EndProcedure
\Procedure{Iterative-Training}{}
\While{Not Converged}
\State Receive a random example $(X, Y^*)$
  \If{Bridge-step}
  \State Draw samples $Y$ from \learner
  \State Update bridge via Equation (\ref{bridge_update})
  \ElsIf{Generator-step}
  \State Draw samples $Y$ from \bridge
  \State Update generator via Equation (\ref{eq:gen_update})
  \EndIf
\EndWhile
\EndProcedure
\end{algorithmic}
\end{algorithm}

\section{Experiment}
We select machine translation and abstractive text summarization as benchmarks to verify our GBN framework. 
\subsection{Similarity Score Function}
\label{sec.smf}
In our experiments, instead of directly using BLEU or ROUGE as reward to guide the bridge network's policy search, we design a simple surrogate n-gram matching reward as follows:
\begin{align}
\small
\begin{split}
S(Y, Y^*) = 0.4N_4 + 0.3N_3 + 0.2N_2 + 0.1N_1
\end{split}
\end{align}
$N_n$ represents the n-gram matching score between $Y$ and $Y^*$. In order to alleviate reward sparsity at sequence level, we further decompose the global reward $S(Y, Y^*)$ as a series of local rewards at every time step. Formally, we write the step-wise reward \rewardt as follows:
\begin{small}
\begin{align}
\label{eq:payoff}
\rewardt &= 
\begin{cases}
1.0; N(y_{1:t}, y_{t-3:t}) \le N(Y^*, y_{t-3:t}) \\
0.6; N(y_{1:t}, y_{t-2:t}) \le N(Y^*, y_{t-2:t}) \\
0.3; N(y_{1:t}, y_{t-1:t}) \le N(Y^*, y_{t-1:t}) \\
0.1; N(y_{1:t}, y_{t}) \le N(Y^*, y_{t}) \\
0.0; otherwise
\end{cases}
\end{align}
\end{small}
where $N(Y, \tilde{Y})$ represents the occurrence of sub-sequence $\tilde{Y}$ in whole sequence $Y$. Specifically, if a certain sub-sequence $y_{t-n+1:t}$ from $Y$ appears less times than in the reference $Y^*$, $y_t$ receives reward. Formally, we rewrite the step-level gradient for each sampled $Y$ as follows:
\begin{align}
\small
\begin{split}
& -\frac{S(Y,Y^*)}{\tau}\nabla \log{\bridge} \\ =& \sum_t -\frac{\rewardt}{\tau} \cdot \nabla \log{\bridget}
\end{split}
\end{align}

\subsection{Machine Translation}
\paragraph{Dataset}
We follow~\citet{ranzato2015sequence,bahdanau2016actor} and select German-English machine translation track of the IWSLT 2014 evaluation campaign. The corpus contains sentence-wise aligned subtitles of TED and TEDx talks. We use Moses toolkit~\cite{koehn2007moses} and remove sentences longer than 50 words as well as lower-casing. 
The evaluation metric is BLEU~\cite{papineni2002bleu} computed via the multi-bleu.perl.

\begin{table}[!t]
\centering
\begin{tabular}{|l|l|l|}
\hline
\textbf{Methods} & \textbf{Baseline} & \textbf{Model} \\
\hline
\hline
MIXER & 20.10 & 21.81 \textcolor{blue}{\small{ +1.71}} \\
\hline
BSO & 24.03 & 26.36 \textcolor{blue}{\small{ +2.33}} \\
\hline
AC & 27.56 & 28.53 \textcolor{blue}{\small{ +0.97}}\\
\hline
Softmax-Q & 27.66 & 28.77 \textcolor{blue}{\small{ +1.11}}\\
\hline
\hline
\makecell{Uniform GBN \\ ($\tau = 0.8$)} & \multirow{8}{*}{29.10} & {29.80\textcolor{blue}{\small{ +0.70}}} \\ 
\cline{1-1} \cline{3-3} \makecell{LM GBN \\ ($\tau = 0.8$)} &  & 29.90\textcolor{blue}{\small{ +0.80}} \\
\cline{1-1} \cline{3-3} \makecell{Coaching GBN \\
($\tau = 0.8$)} &  & 29.98\textcolor{blue}{\small{ +0.88}}\\
\cline{1-1} \cline{3-3} \makecell{Coaching GBN \\
($\tau = 1.2$)} &  & 30.15\textcolor{blue}{\small{ +1.05}} \\
\cline{1-1} \cline{3-3} \makecell{Coaching GBN \\
($\tau = 1.0$)} &  & \textbf{30.18}\textcolor{blue}{\small{ +1.08}}\\
\hline
\end{tabular}
\caption{Comparison with existing works on IWSLT-2014 German-English Machine Translation Task.}
\label{tab:results-de2en}
\end{table}

\begin{figure}[!t]
\centering
\includegraphics[width=0.9\linewidth]{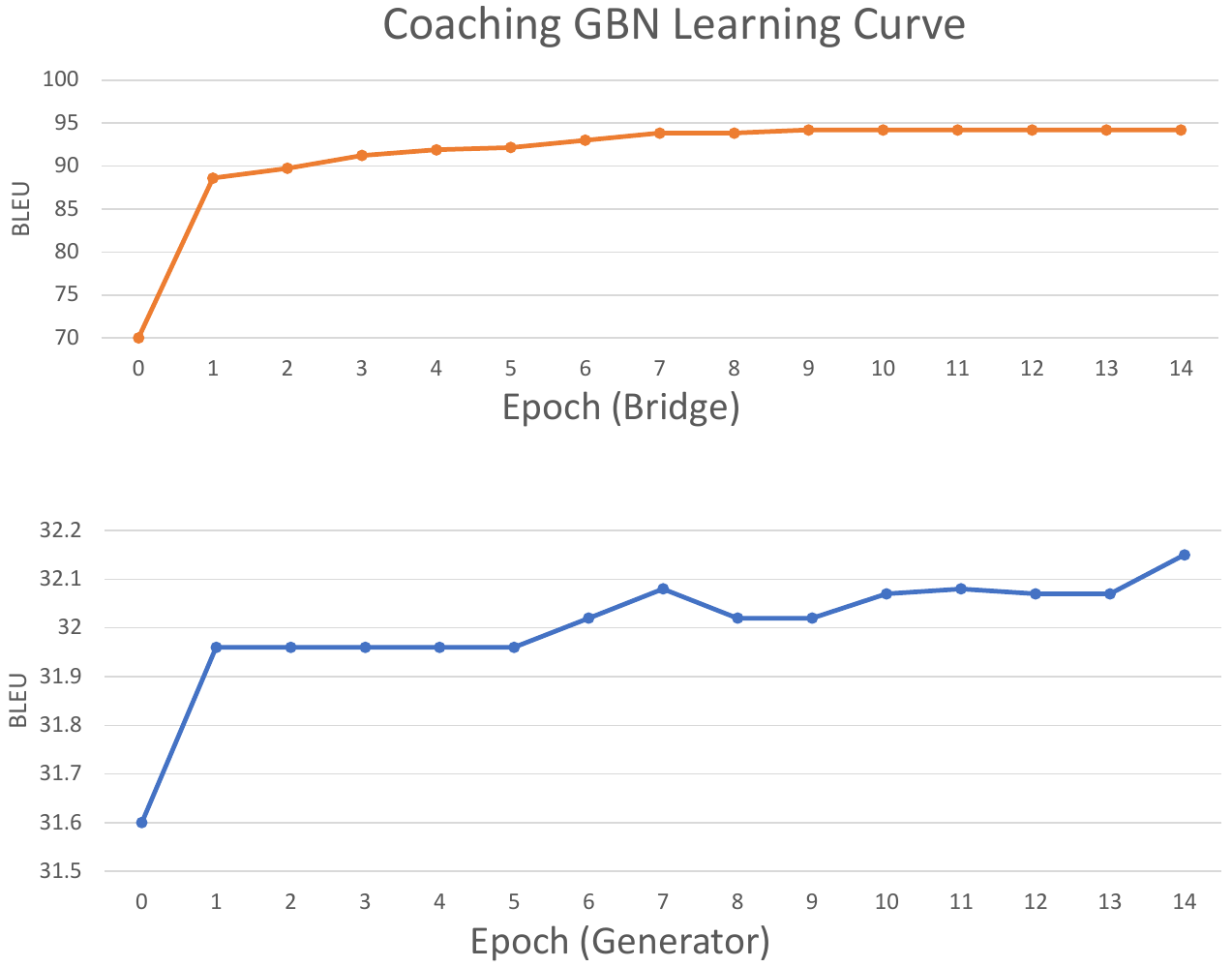}
\caption{Coaching GBN's learning curve on IWSLT German-English Dev set.}
\label{fig:learning-curve}
\end{figure}

\paragraph{System Setting}
We use a unified GRU-based RNN~\cite{chung2014empirical} for both the generator and the coaching bridge. In order to compare with existing papers, we use a similar system setting with 512 RNN hidden units and 256 as embedding size. We use attentive encoder-decoder to build our system~\cite{bahdanau2014neural}. During training, we apply ADADELTA~\cite{zeiler2012adadelta} with $\epsilon=10^{-6}$ and $\rho=0.95$ to optimize parameters of the generator and the coaching bridge. During decoding, a beam size of 8 is used to approximate the full search space. An important hyper-parameter for our experiments is the temperature $\tau$. For the uniform/LM bridge, we follow~\citet{norouzi2016reward} to adopt an optimal temperature $\tau=0.8$. And for the coaching bridge, we test hyper-parameters from $\tau \in {\{0.8, 1.0, 1.2\}}$. Besides comparing with our fine-tuned baseline, other systems for comparison of relative BLEU improvement are: MIXER~\cite{ranzato2015sequence}, BSO~\cite{wiseman2016sequence}, AC~\cite{bahdanau2016actor}, Softmax-Q~\cite{ma2017softmax}.

\paragraph{Results}
The experimental results are summarized in Table~\ref{tab:results-de2en}. We can observe that our fine-tuned MLE baseline (29.10) is already over-competing other systems and our proposed GBN can yield a further improvement. We also observe that LM GBN and coaching GBN have both achieved better performance than Uniform GBN, which confirms that better regularization effects are achieved, and the generators become more robust and generalize better.
We draw the learning curve of both the bridge and the generator in Figure~\ref{fig:learning-curve} to demonstrate how they cooperate during training. We can easily observe the interaction between them: as the generator makes progress, the coaching bridge also improves itself to propose harsher targets for the generator to learn.

\subsection{Abstractive Text Summarization}
\paragraph{Dataset} 
We follow the previous works by~\citet{rush2015neural,zhou2017selective} and use the same corpus from Annotated English Gigaword dataset~\cite{napoles2012annotated}. In order to be comparable, we use the same script~\footnote{https://github.com/facebookarchive/NAMAS} released by~\citet{rush2015neural} to pre-process and extract the training and validation sets. For the test set, we use the English Gigaword, released by~\citet{rush2015neural}, and evaluate our system through ROUGE~\cite{lin2004rouge}. Following previous works, we employ ROUGE-1, ROUGE-2, and ROUGE-L as the evaluation metrics in the reported experimental results.

\begin{table}[!t]
\centering
\begin{tabular}{|l|l|l|l|}
\hline
\textbf{Methods} & \textbf{RG-1} & \textbf{RG-2} & \textbf{RG-L} \\
\hline
\hline
ABS & 29.55 & 11.32 & 26.42 \\
\hline 
ABS+ & 29.76 & 11.88 & 26.96 \\
\hline
Luong-NMT & 33.10 & 14.45 & 30.71\\
\hline
SAEASS & \textbf{36.15} & \textbf{17.54} & \textbf{33.63}\\
\hline
\hline
seq2seq+att & 34.04 & 15.95 & 31.68 \\
\hline
\makecell{Uniform GBN \\($\tau = 0.8$)} & 34.10 & 16.70 & 31.75 \\
\hline
\makecell{LM GBN \\($\tau = 0.8$)} & 34.32 & 16.88 & 31.89 \\
\hline
\makecell{Coaching GBN \\($\tau = 0.8$)} & 34.49  & 16.70 & 31.95\\
\hline
\makecell{Coaching GBN \\($\tau = 1.2$)} & 34.83 & 16.83 & 32.25 \\
\hline
\makecell{Coaching GBN \\($\tau = 1.0$)} & \textbf{35.26} & \textbf{17.22} & \textbf{32.67} \\
\hline
\end{tabular}
\caption{Full length ROUGE F1 evaluation results on the English Gigaword test set used by ~\cite{rush2015neural}. RG in the Table denotes ROUGE. Results for comparison are taken from SAEASS~\cite{zhou2017selective}.}
\label{tab:results-abs}
\end{table}

\paragraph{System Setting}
We follow~\citet{zhou2017selective,rush2015neural} to set input and output vocabularies to 119,504 and 68,883 respectively, and we also set the word embedding size to 300 and all GRU hidden state size to 512. Then we adopt dropout~\cite{srivastava2014dropout} with probability $p = 0.5$ strategy in our output layer. We use attention-based sequence-to-sequence model~\cite{bahdanau2014neural,cho2014learning} as our baseline and reproduce the results of the baseline reported in~\citet{zhou2017selective}. As stated, the attentive encoder-decode architecture can already outperform existing ABS/ABS+ systems~\cite{rush2015neural}. In coaching GBN, due to the fact that the input of abstractive summarization $X$ contains more information than the summary target $Y^*$, directly training the bridge \bridge to understand the generator \learner is infeasible. Therefore, we re-design the coaching bridge to receive both source and target input $X, Y$ and we enlarge its vocabulary size to 88,883 to encompass more information about the source side. In Uniform/LM GBN experiments, we also fix the hyper-parameter $\tau=0.8$ as the optimal setting.

\begin{figure}[!t]
\centering
\includegraphics[width=0.9\linewidth]{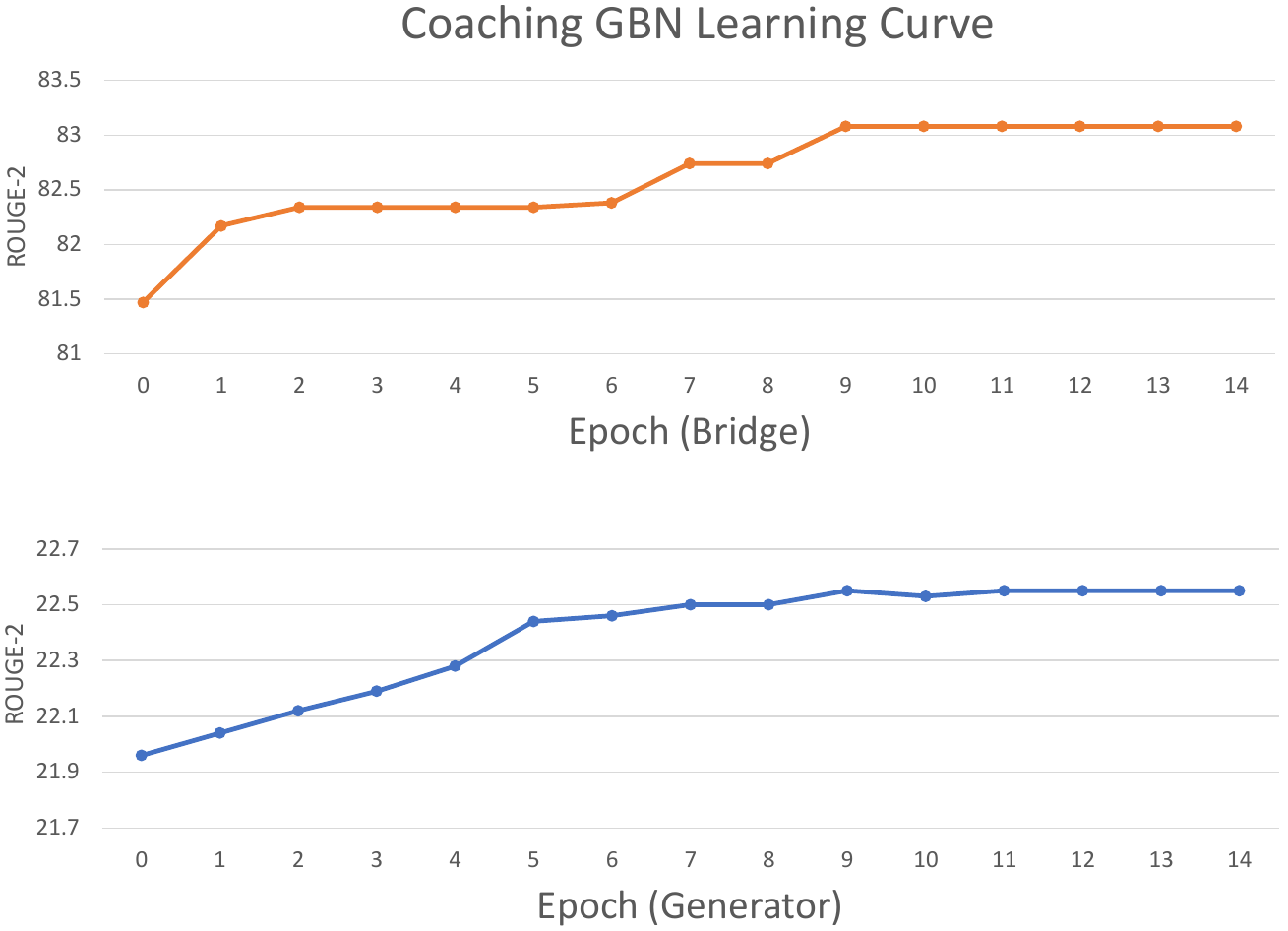}
\caption{Coaching GBN's learning curve on Abstractive Text Summarization Dev set.}
\label{fig:learning-curve-sum}
\end{figure}

\paragraph{Results}
The experimental results are summarized in Table ~\ref{tab:results-abs}. We can observe a significant improvement via our GBN systems. Similarly, the coaching GBN system achieves the strongest performance among all, which again reflects our assumption that more sophisticated regularization can benefit generator's training. We draw the learning curve of the coaching GBN in Figure~\ref{fig:learning-curve-sum} to demonstrate how the bridge and the generator promote each other.

\section{Analysis}
By introducing different constraints into the bridge module, the bridge distribution will propose different training samples for the generator to learn. From Table~\ref{tab:samples}, we can observe that most samples still reserve their original meaning. The uniform bridge simply performs random replacement without considering any linguistic constraint. The LM bridge strives to smooth reference sentence with high-frequent words. And the coaching bridge simplifies difficult expressions to relieve generator's learning burden. From our experimental results, the more rational and aggressive diversification from the coaching GBN clearly benefits generator the most and helps the generator generalize to more unseen scenarios.

\section{Related Literature}

\subsection{Data Augmentation and Self-training}
In order to resolve the data sparsity problem in Neural Machine Translation (NMT), many works have been conducted to augment the dataset. The most popular strategy is via self-learning, which incorporates the self-generated data directly into training. ~\citet{zhang2016exploiting} and ~\citet{sennrich2015improving} both use self-learning to leverage massive monolingual data for NMT training. 
Our bridge can take advantage of the parallel training data only, instead of external monolingual ones to synthesize new training data.

\subsection{Reward Augmented Maximum Likelihood}
Reward augmented maximum likelihood or RAML~\cite{norouzi2016reward} proposes to integrate task-level reward into MLE training by using an exponentiated payoff distribution. KL divergence between the payoff distribution and the generator's output distribution are minimized to achieve an optimal task-level reward. 
Following this work, ~\citet{ma2017softmax} introduces softmax Q-Distribution to interpret RAML and reveals its relation with Bayesian decision theory. These two works both alleviate data sparsity problem by augmenting target examples based on the ground truth. Our method draws inspiration from them but seeks to propose the more general Generative Bridging Network, which can transform the ground truth into different bridge distributions, from where samples are drawn will account for different interpretable factors.

\subsection{Coaching}
Our coaching GBN system is inspired by imitation learning by coaching~\cite{he2012imitation}. 
Instead of directly behavior cloning the oracle, they advocate learning hope actions as targets from a coach which is interpolated between learner's policy and the environment loss. As the learner makes progress, the targets provided by the coach will become harsher to gradually improve the learner. Similarly, our proposed coaching GBN is motivated to construct an easy-to-learn bridge distribution which lies in between the ground truth and the generator. Our experimental results confirm its effectiveness to relieve the learning burden.

\begin{table}[!t]
\begin{tabular}{|l|l|}
\hline
\textbf{System} & \textbf{Uniform GBN} \\
\hline
Property & Random Replacement \\
\hline
Reference & \small{the question \textcolor{blue}{is} , is it worth it ?}\\
\hline
Bridge & \small{the question \textcolor{blue}{lemon} , was it worth it ?}\\
\hline
\hline
\textbf{System} & \textbf{Language-model GBN} \\
\hline
Property & Word Replacement \\
\hline
Reference & \small{\textcolor{blue}{now} how can this help us ?} \\
\hline
Bridge & \small{\textcolor{blue}{so} how can this help us ?} \\
\hline
\hline
\textbf{System} & \textbf{Coaching GBN} \\
\hline
Property & Reordering \\
\hline
Reference & \small{i need \textcolor{blue}{to have a health care lexicon .}} \\
\hline
Bridge & \small{i need \textcolor{blue}{a lexicon for health care .}}\\
\hline
Property & Simplification\\
\hline
Reference & \tabincell{l}{\small{\textcolor{blue}{this is the way that} most of us were taught} \\ \small{to \textcolor{blue}{tie} our shoes .}}\\
\hline
Bridge & \small{most of us learned to \textcolor{blue}{bind} our shoes .}\\
\hline
\end{tabular}
\caption{Qualitative analysis for three different bridge distributions.}
\label{tab:samples}
\end{table}


\section{Conclusion}
In this paper, we present the Generative Bridging Network (GBN) to overcome data sparsity and overfitting issues with Maximum Likelihood Estimation in neural sequence prediction. Our implemented systems prove to significantly improve the performance, compared with strong baselines. We believe the concept of bridge distribution can be applicable to a wide range of distribution matching tasks in probabilistic learning. 
In the future, we intend to explore more about GBN's applications as well as its provable computational and statistical guarantees. 

\bibliography{naaclhlt2018}
\bibliographystyle{acl_natbib}

\clearpage
\appendix

\section{Supplemental Material}
This part first provides detailed derivation of Equation (\ref{uniform_bridge_obj_mod}) and (\ref{lm_bridge_obj_mod}) from Equation (\ref{uniform_bridge_obj}) and (\ref{lm_bridge_obj}), since our uniform bridge distribution and language-model bridge distribution have closed-form solutions given a fixed uniform distribution and a language model as constraints. Then, we give explanation of Equation (\ref{coaching_bridge_obj}), the objective function of coaching bridge, where the constraint is the inverse KL compared with previous two bridges and then give detailed derivation of the gradient update Equation (\ref{bridge_update}). 

\paragraph{Derivation of Equation (\ref{uniform_bridge_obj_mod})}

\begin{align}
\begin{split}
&\lossb \\
=& \expect{Y \sim p_\eta}  -\similaritytau + KL(\bridge \vert \vert U(Y)) \\
=& \int_Y -\bridge \log \exp(\similaritytau) \\
&+ \int_Y \bridge \log \frac{\bridge}{U(Y)} \\
=& \int_Y \bridge \log \frac{\bridge}{\exp(\similaritytau) \cdot U(Y)} \\
=& \int_Y \bridge \log \frac{\bridge}{\exp(\similaritytau) \cdot \frac{1}{\vert \mathcal{Y} \vert}} \\
=& \int_Y \bridge \log \frac{\bridge}{\exp{\similaritytau}} \\
&+ \log {\vert \mathcal{Y} \vert} \int_Y \bridge \\
=& \int_Y \bridge \log \frac{\bridge}{\exp{\similaritytau}} + Const \\
=& \int_Y \bridge \log \frac{\bridge}{\frac{\exp \similaritytau}{Z}} + Const' \\
&= KL(\bridge \vert \vert \frac{\exp \similaritytau}{Z}) + Const'
\end{split}
\end{align}

Here, the $Y^*$ related constant $Z$ is needed to transform a unnormalized similarity score to a probability:
\begin{equation}
Z(Y^*) = \int_Y \exp \similaritytau
\end{equation}
 
\paragraph{Derivation of Equation (\ref{lm_bridge_obj_mod})}

\begin{align}
\begin{split}
&\lossb \\
=& \expect{Y \sim p_\eta}  -\similaritytau + KL(\bridge \vert \vert p_{LM}(Y)) \\
=& \int_Y -\bridge \log \exp(\similaritytau) \\
&+ \int_Y \bridge \log \frac{\bridge}{p_{LM}(Y)} \\
=& \int_Y \bridge \log \frac{\bridge}{\exp(\similaritytau) \cdot p_{LM}(Y)} \\
=& \int_Y \bridge \log \frac{\bridge}{\frac{\exp{\similaritytau} \cdot P_{LM}(Y)}{Z}} + Const \\
=& KL(\bridge \vert \vert \frac{\exp \similaritytau \cdot P_{LM}(Y)}{Z}) + Const'
\end{split}
\end{align}

Here, the $Y^*$ related constant $Z$ is needed to transform a unnormalized \textit{weighted} similarity score to a probability:
\begin{equation}
Z(Y^*) = \int_Y \exp \similaritytau \cdot P_{LM}(Y)
\end{equation}

\paragraph{Explanation of Equation (\ref{coaching_bridge_obj})} 

This equation is the objective function of our coaching bridge, which uses an inverse KL term\footnote{That is the use of $KL(p_\theta \vert \vert p_\eta)$ instead of $KL(p_\eta \vert \vert p_\theta)$.} as part of its objective. The use of inverse KL is out of the consideration of computational stability. The reasons are two-fold: 1). the inverse KL will do not change the effect of the constraint; 2). the inverse KL requires sampling from the generator and uses those samples as the target to train the bridge, which has the same gradient update ad MLE, so we do not need to consider baseline tricks in Reinforcement Learning implementation. 

\paragraph{Gradient derivation of Equation (\ref{coaching_bridge_obj})}

\begin{align}
\begin{split}
&\nabla \lossb \\
=& \nabla_\eta \expect{Y \sim \bridge} -\similaritytau + \nabla_\eta KL(\learner \vert \vert \bridge) \\
=& \expect{Y \sim \bridge} -\similaritytau \nabla_\eta \log \bridge \\
&+ \nabla_\eta \expect{Y \sim \learner} \log \bridge \\
=& \expect{Y \sim \bridge} -\similaritytau \nabla \log \bridge \\
&+ \expect{Y \sim \learner} \nabla \log\bridge
\end{split}
\end{align} 

\end{document}